\documentclass[twocolumn]{article}
\usepackage{graphicx}
\usepackage{amsmath}
\usepackage{hyperref}

\usepackage{pgfplots}
\pgfplotsset{compat=1.18}

\usepackage{makecell}

\usepackage{rotating}
\usepackage{booktabs}

\usepackage{soul}
\usepackage{xcolor}

\usepackage{subcaption}

\definecolor{myyellow}{RGB}{255,255,200}  % Softer yellow
\sethlcolor{myyellow}

\usepackage{float}
\usepackage[margin=1in]{geometry}
\usepackage{listings}
\usepackage[
  style=ieee,   % <- IEEE bibliography + citation style
  sorting=none  % keep your order
]{biblatex}
\usepackage[compact]{titlesec}

\usepackage[none]{hyphenat}

\usepackage{tabularx}
\usepackage{array} % for >{\raggedright\arraybackslash}

\usepackage{tikz}
\usetikzlibrary{positioning, arrows.meta, shapes.geometric, shapes.misc, calc}

\usepackage{booktabs}
\usepackage{tabularx}
\usepackage{pifont}
\newcommand{\checkmark}{\ding{51}}

\usepackage{listings}
\title{ALIVE: An Avatar-Lecture Interactive Video Engine with Content-Aware Retrieval for Real-Time Interaction}

\author{
  Md Zabirul Islam  \\
  Department of Computer Science \\
  Rensselaer Polytechnic Institute \\
  Troy, New York 12180, USA
  \and
  Md Motaleb Hossen Manik \\
  Department of Computer Science \\
  Rensselaer Polytechnic Institute \\
  Troy, New York 12180, USA
  \and
  Ge Wang  \thanks{Corresponding author: Ge Wang, email: wangg6@rpi.edu}\\
  Department of Biomedical Engineering  \\
  Rensselaer Polytechnic Institute \\
  Troy, New York 12180, USA
}

\date{}

\begin{document}

\maketitle

\begin{abstract}
Traditional lecture videos offer flexibility but lack mechanisms for real-time clarification, forcing learners to search externally when confusion arises. Recent advances in large language models and neural avatars provide new opportunities for interactive learning, yet existing systems typically lack lecture awareness, rely on cloud-based services, or fail to integrate retrieval and avatar-delivered explanations in a unified, privacy-preserving pipeline.

We present \textbf{ALIVE}, an \textbf{A}vatar-\textbf{L}ecture \textbf{I}nteractive \textbf{V}ideo \textbf{E}ngine that transforms passive lecture viewing into a dynamic, real-time learning experience. ALIVE operates fully on local hardware and integrates (1) Avatar-delivered lecture generated through ASR transcription, LLM refinement, and neural talking-head synthesis; (2) A content-aware retrieval mechanism that combines semantic similarity with timestamp alignment to surface contextually relevant lecture segments; and (3) Real-time multimodal interaction, enabling students to pause the lecture, ask questions through text or voice, and receive grounded explanations either as text or as avatar-delivered responses.

To maintain responsiveness, ALIVE employs lightweight embedding models, FAISS-based retrieval, and segmented avatar synthesis with progressive preloading. We demonstrate the system on a complete medical imaging course, evaluate its retrieval accuracy, latency characteristics, and user experience, and show that ALIVE provides accurate, content-aware, and engaging real-time support.

ALIVE illustrates how multimodal AI—when combined with content-aware retrieval and local deployment—can significantly enhance the pedagogical value of recorded lectures, offering an extensible pathway toward next-generation interactive learning environments.

\end{abstract}

\noindent\textbf{Keywords:}
Interactive Lecture Systems,
Content-Aware Retrieval,
Neural Talking-Head Avatars,
Multimodal AI,
Privacy-Preserving Learning

\section{Introduction}

The rapid expansion of digital education has transformed how instructional content is delivered and consumed. Recorded lectures have become a cornerstone of online learning, offering flexibility, repeatability, and broad access to high-quality instruction independent of time and location. Despite these advantages, most lecture videos remain fundamentally passive: when students encounter confusion, they must pause the lecture and seek clarification through external resources, often without guidance on relevance, accuracy, or alignment with the instructor’s intent \cite{baker2022bias}. The absence of timely, context-aware feedback continues to limit the pedagogical effectiveness of otherwise valuable recorded lectures.

These limitations are especially pronounced in technically demanding domains such as medical imaging, where learning depends on precise reasoning about physical principles, mathematical models, and algorithmic pipelines. Even carefully designed lectures cannot anticipate every point of difficulty. When students struggle to interpret an imaging artifact, follow a reconstruction workflow, or connect concepts across modalities, the lack of real-time, content-aware support disrupts learning continuity and increases cognitive load. In such settings, effective clarification must be both contextually grounded in the instructional material and temporally aligned with the moment confusion arises.

Recent advances in large language models (LLMs) have enabled conversational systems capable of generating fluent, human-like explanations and supporting open-ended inquiry \cite{achiam2023gpt,touvron2023llama,team2023gemini,openai2023chatgpt}. When combined with retrieval-augmented generation (RAG), these models can ground responses in external instructional content, improving factual accuracy and domain relevance \cite{lewis2020retrieval}. In parallel, research on pedagogical avatars and animated instructors suggests that human-like visual presentation can enhance the clarity and continuity of instructional delivery \cite{anasingaraju2016digital,johnson2016face,fink2024ai}.

Despite these advances, existing educational systems typically employ these components in isolation. Most LLM-based tools operate as standalone chat interfaces, disconnected from the lecture timeline and unaware of the instructional context in which a question arises. Avatar-delivered  lecture systems often rely on fixed scripts or limited interaction, lacking real-time reasoning and retrieval capabilities. Furthermore, many deployed solutions depend on cloud-based services, raising concerns about privacy, reproducibility, and deployment in resource-constrained or sensitive educational environments. As a result, prior systems rarely provide a unified solution that combines lecture awareness, content-aware reasoning, real-time interaction, and avatar-delivered 
explanation within a fully local pipeline.

To address these challenges, we introduce \textbf{ALIVE}, a fully local multimodal system that transforms passive lecture videos into interactive, context-aware learning experiences. ALIVE integrates avatar-delivered lecture delivery with real-time, content-aware question answering. Students can pause a lecture at any moment, ask questions using text or voice, and receive explanations that are grounded in the most relevant lecture segments. Responses are delivered either as text or as avatar-delivered explanations, preserving instructional continuity while enabling immediate clarification.

A central component of ALIVE is its \emph{content-aware retrieval} mechanism, which combines semantic similarity with timestamp alignment to identify lecture segments that are both topically relevant and temporally aligned with the moment of inquiry. This design ensures that generated explanations remain consistent with the instructor’s intent and the surrounding instructional context. To support responsive interaction, ALIVE employs lightweight retrieval models and segmented avatar synthesis with progressive preloading, reducing perceived latency while maintaining visual continuity during avatar-delivered responses.

We demonstrate ALIVE on a complete medical imaging course taught by Professor Ge Wang, covering foundational topics such as X-ray computed tomography, magnetic resonance imaging, and tomographic reconstruction. Through quantitative evaluation of retrieval behavior and system latency, we show that ALIVE provides accurate, content-aware, and responsive real-time support during lecture viewing under fully local deployment.

% This work makes the following contributions:
% \begin{itemize}
%     \item \textbf{Fully local multimodal interactive lecture engine.}  
%     We design ALIVE as an end-to-end system in which all components—including speech recognition, retrieval, LLM inference, text-to-speech, and avatar synthesis—run locally, enabling privacy preservation, reproducibility, and deployment without reliance on cloud services.

%     \item \textbf{Content-aware retrieval for  question answering.}  
%     We introduce a retrieval mechanism that combines semantic similarity with timestamp alignment, allowing student questions to be grounded in lecture segments that are both topically relevant and temporally aligned with the moment of inquiry.

%     \item \textbf{Real-time multimodal interaction during lecture playback.}  
%     ALIVE enables students to pause a lecture and ask questions using text or voice, providing immediate, context-specific explanations without disrupting the natural flow of learning.

%     \item \textbf{Segmented avatar-delivered responses with low perceived latency.}  
%     We integrate offline text-to-speech with neural talking-head synthesis and segmented video generation, allowing avatar-delivered explanations to begin quickly and play smoothly even for longer responses.
% \end{itemize}

This work makes the following contributions:
\begin{itemize}
    \item \textbf{Fully local multimodal interactive lecture engine.}  
    We design ALIVE as an end-to-end system in which all components—including speech recognition, content-aware retrieval, LLM inference, text-to-speech, and avatar synthesis—run entirely on local hardware, enabling privacy-preserving deployment, reproducible experimentation, and use in educational settings where cloud-based services are impractical or undesirable.

    \item \textbf{Content-aware retrieval for question answering.}  
    We introduce a retrieval mechanism that combines semantic similarity with timestamp alignment, ensuring that student questions are grounded in lecture segments that are both topically relevant and temporally aligned with the moment of inquiry, thereby preserving instructional intent and contextual consistency.

    \item \textbf{Real-time multimodal interaction during lecture playback.}  
    ALIVE enables students to pause a lecture and ask questions using text or voice, providing immediate, context-specific explanations that support clarification at the point of confusion without disrupting the natural flow of lecture viewing.

    \item \textbf{Segmented avatar-delivered responses with low perceived latency.}  
    We integrate offline text-to-speech with neural talking-head synthesis and segmented video generation, allowing avatar-delivered explanations to begin promptly and play smoothly even for longer responses, which reduces perceived latency while maintaining an engaging, instructor-style presentation.
\end{itemize}

The remainder of this paper details the design, implementation, and system-level evaluation of ALIVE. Section~\ref{sec:overview} presents a system overview. Section~\ref{sec:lectureprep} describes the lecture preparation pipeline. Section~\ref{sec:qa} and Section~\ref{sec:avatar} detail the interactive question answering and avatar-delivered response generation modules, respectively. Section~\ref{sec:frontend} discusses the frontend design, followed by evaluation in Section~\ref{sec:evaluation}. Section~\ref{sec:discussion} analyzes limitations and implications, and Section~\ref{sec:conclusion} concludes the paper.

\section{Related Works}
Digital humans—highly realistic virtual representations of people—have been increasingly adopted in areas such as customer service, entertainment, and broadcasting. In educational contexts, these systems are often described as pedagogical avatars or pedagogical agents. Early studies showed that virtual instructors could improve learner engagement, but these systems were generally limited by low realism and fixed, preprogrammed knowledge bases \cite{graesser2005autotutor}. Advances in artificial intelligence have begun to address these limitations. LLMs now provide broader and more flexible knowledge sources, while modern speech-driven animation techniques enable more natural synchronization of lip movements, facial expressions, and conversational cues. These developments extend the capabilities of avatar-delivered learning environments and motivate systems that support richer, real-time instructional interactions.

Building on these advances, interactive learning systems have emerged at the intersection of several rapidly evolving areas, including avatar-delivered  educational interfaces, neural talking-head synthesis, video question answering (Video~QA), RAG, and research on avatar realism. Prior work has examined how virtual instructors influence learner engagement, how neural networks generate photorealistic talking-head avatars, how long video sequences can be searched efficiently, and how retrieval techniques can improve the factual grounding of LLM-generated answers. Together, these research directions form the foundation for ALIVE and highlight the need for systems that integrate avatar-delivered presentation, time-aware retrieval, and LLM-driven explanation within a single content-aware framework.

In this section, we review the most relevant literature across five categories: (1) virtual avatars in education, (2) neural talking-head generation methods, (3) Video~QA and video-based RAG frameworks, (4) retrieval-augmented educational question answering, and (5) recent advances in talking-head avatar synthesis and perception.

\subsection{Virtual Avatars for Educational Engagement}

A growing body of work has examined how virtual avatars influence learning effectiveness, engagement, and user experience. Recent studies show that avatar-delivered educational videos can enhance learners’ emotional experience, engagement, and perceived learning outcomes. For example, Zhang and Wu \cite{zhang2024impact} conducted a structural-equation–modeling study demonstrating that both video quality and avatar expressiveness improve learners’ emotional responses and engagement with instructional content. Their results further indicate that avatar expressiveness—such as natural movement, facial expressions, and visual appeal—positively affects learning effectiveness.

Research on educational robotics and embodied agents similarly suggests that anthropomorphic avatars can strengthen social presence and support better instructional outcomes. Kodani et al. \cite{kodani2025android} reported that android-style avatars increase learners’ cognitive engagement by eliciting embodied anthropomorphization, which enhances attention and improves learning performance compared to non-embodied presentation formats.

Another line of work investigates how speech intelligibility and synchronized visual motion contribute to learner comprehension. Cioffi et al. \cite{cioffi2025speech} found that audio–visual facial animation significantly improves perceived intelligibility over audio-only explanations, underscoring the pedagogical value of synchronized lip movements when delivering spoken instructional content through avatars.

\subsection{Neural Talking-Head Avatars}

Recent advances in neural head-avatar synthesis have made high-quality talking-head generation feasible with minimal input data. Zakharov et al. \cite{zakharov2020fast} introduced a bi-layer neural synthesis framework capable of producing realistic talking-head avatars from a single portrait image. Their method generates temporally coherent facial motion driven by audio or other control signals. Building on this direction, Grassal et al. \cite{grassal2022neural} proposed a monocular RGB-based approach that can reproduce fine-grained head pose and expression dynamics with high fidelity.

These techniques provide the foundation for many modern talking-head systems by supporting realistic facial motion, accurate lip synchronization, and consistent identity preservation. Such capabilities are essential for systems like ALIVE, which rely on neural avatar animation to deliver interactive, context-aware lecture responses.

\subsection{Video QA and Video--RAG Systems}

Video~QA typically requires deep temporal reasoning, multimodal fusion, and large annotated datasets. Early approaches treated Video~QA as a supervised learning task that depended on densely labeled video--question--answer pairs \cite{su2021end}. However, the cost of annotation and the difficulty of modeling long-range temporal dependencies have led to increased interest in retrieval-augmented and generative methods.

A representative supervised method is the Generator--Pretester Network proposed by Su et~al.\ \cite{su2021end}, which jointly generates video-conditioned question--answer pairs and evaluates them using a pretesting module. Their framework incorporates appearance and object features to improve QA accuracy and demonstrates that automatically generated question--answer pairs can benefit downstream training. Despite these contributions, supervised Video~QA systems remain computationally demanding and reliant on large-scale labeled datasets.

To address these limitations, recent retrieval-augmented frameworks index video segments and retrieve only the most relevant portions during inference. Xu et~al.\ \cite{xu2023retrieval} introduced a retrieval-based Video Language Model that embeds video segments and uses similarity search to support efficient long-video QA, reducing memory usage by avoiding full-sequence encoding. Ren et~al.\ extended this approach with VideoRAG \cite{ren2025videoragretrievalaugmentedgenerationextreme}, which applies retrieval-augmented generation to extremely long-context videos through multi-stage retrieval and finer-grained temporal selection.

Together, these studies highlight two major trends: (1) retrieval as a scalable mechanism for grounding model outputs in long-duration audiovisual content, and (2) generative models that synthesize answers from retrieved segments. ALIVE follows these trends but is designed for a distinct educational context in which \emph{timestamp-aware retrieval} is essential. 
% Unlike prior systems built for open-domain video content, ALIVE uses subtitle-based semantic indexing and pause-aligned retrieval to provide grounded, context-specific explanations that correspond directly to the instructional moment in which a student asks a question.
Unlike prior systems designed for open-domain video content, ALIVE uses subtitle-based semantic indexing and pause-aligned retrieval to deliver grounded, context-specific explanations tied directly to the moment a student asks a question.

\subsection{Retrieval-Augmented Educational Question Answering}

LLMs often struggle with factual grounding when answering domain-specific questions, which motivates the use of RAG to incorporate external knowledge sources. Recent work in educational question answering (QA) has increasingly emphasized multimodal retrieval, discipline-aware indexing strategies, and stronger grounding mechanisms.

MDKAG \cite{zhao2025mdkag} introduces a multimodal disciplinary knowledge graph that integrates text, images, and structured metadata for course-level QA. By combining graph-based retrieval with an LLM generator, the method improves factual consistency and contextual relevance in educational settings. Unlike purely text-based retrieval pipelines, MDKAG explicitly models domain relationships, enabling more semantically aligned evidence retrieval.

In related work, Alawwad et al.~\cite{alawwad2025enhancing} examine how RAG improves LLM performance on textbook question answering. Their results show that even strong LLMs benefit from retrieval, particularly for questions requiring multi-step reasoning or conceptual linkage. The study also highlights ongoing challenges in selecting retrieval granularity that matches question difficulty—an issue directly relevant to interactive lecture systems.

For long-form and multimodal educational content such as lecture recordings, adaptive retrieval becomes even more critical. AdaVideoRAG \cite{xue2025adavideorag} proposes an omni-contextual retrieval mechanism that dynamically selects evidence across spatial, temporal, and semantic dimensions for long-video understanding. Although designed for general video analysis, this work demonstrates the importance of multi-scale and adaptive retrieval strategies, principles that we similarly adopt for time-aware transcript retrieval.

Together, these studies illustrate a broader shift from simple keyword search toward structured, multimodal, and context-adaptive retrieval pipelines. Our system builds on these insights by implementing a fully local, time-indexed retrieval architecture tailored to long-duration lecture videos, enabling grounded and context-specific student interaction.

\subsection{Talking-Head Generation and Neural Avatar Synthesis}

Talking-head generation has advanced rapidly with the development of deep generative models, enabling photorealistic reenactment and controllable avatar synthesis. A recent comprehensive survey by Rakesh et al.~\cite{rakesh2025advancingtalkingheadgeneration} provides an extensive taxonomy of multimodal approaches—including 2D, 3D, NeRF-based, and diffusion frameworks—and highlights ongoing challenges related to identity preservation, temporal consistency, and expressive control. Their analysis underscores the growing relevance of talking-head models for education, telepresence, and interactive learning applications.

Diffusion-based methods have recently improved both realism and stability. ConsistentAvatar~\cite{yang2024consistentavatar} introduces a diffusion framework guided by a temporally sensitive detail representation to jointly enhance temporal coherence, 3D consistency, and expression fidelity. By explicitly modeling temporal patterns, the approach reduces frame-to-frame drift, a key issue for conversational avatar responses.

The perceptual realism of educational avatars has also been studied empirically. Oyekoya and Baffour~\cite{oyekoya2025perception} examine how factors such as head shape, texture fidelity, and avatar orientation influence student perception in instructor-style avatars. Their findings highlight the importance of consistent geometry and naturalistic rendering for maintaining learner engagement—an insight that directly informs the design of avatar front-ends in interactive learning systems.

Together, these works demonstrate substantial progress in neural avatar generation while identifying remaining challenges in real-time rendering, expression control, and reliability. Our system builds on these insights by employing a SadTalker-based pipeline for offline talking-head synthesis, optimized for responsiveness through short-segment rendering and progressive preloading. Throughout this paper, ‘offline’ refers to local, non-cloud execution, not necessarily pre-computation.

\begin{figure*}[t]
    \centering
    \includegraphics[width=0.7\linewidth]{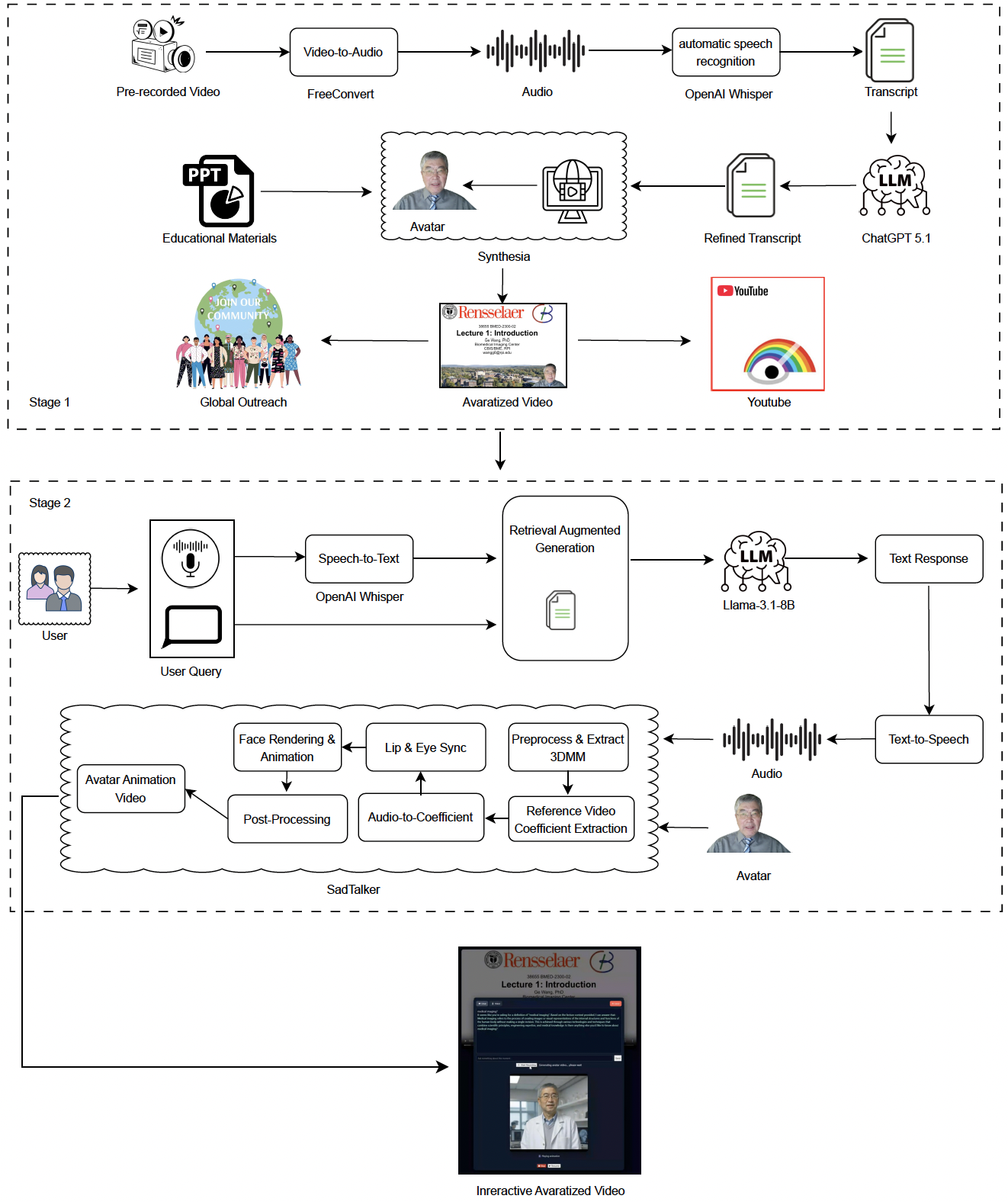}
    % \caption{System overview of ALIVE, the proposed multimodal interactive lecture system. 
    % The frontend video player triggers the question–answering interface, while the backend performs 
    % ASR, time-aware retrieval, LLM inference, text-to-speech, and neural avatar synthesis.}
\caption{System overview of ALIVE, illustrating the fully local, content-aware retrieval and segmented avatar synthesis pipeline that enables real-time interactive lecture engagement.}
    \label{fig:overview}
\end{figure*}

\section{System Overview}
\label{sec:overview}

ALIVE transforms conventional lecture videos into an interactive, avatar-augmented learning environment that operates entirely on local hardware. The system is designed around three tightly integrated stages: (1) Avatar-delivered lecture preparation, (2) Interactive question answering, and (3) Avatar-delivered response generation. Together, these stages enable real-time, content-aware interaction during lecture playback while preserving instructor presence and ensuring user privacy. A high-level illustration of the overall architecture is shown in Figure~\ref{fig:overview}.

\subsection{Architecture Summary}

At a high level, ALIVE integrates a lecture video player, a content-aware retrieval and reasoning backend, and an avatar-delivered response module within a unified web application. The frontend runs in the student’s browser and manages lecture playback, pause detection, multimodal query input (text or voice), and the presentation of textual or avatar-delivered responses. The backend performs all reasoning and generation tasks required to support interactive question answering and avatar synthesis.

A central design principle of ALIVE is \emph{fully local execution}. All processing—including speech recognition, retrieval, language model inference, text-to-speech synthesis, and avatar animation—is performed on the user’s machine. This design ensures that lecture content, student queries, and generated responses remain private, while also enabling reproducible experimentation and deployment in institutional environments without reliance on cloud-based services.

\subsection{Stage 1: Avatar-Delivered Lecture Preparation}

The first stage converts raw lecture recordings into an avatar-delivered video series and constructs a retrieval-ready representation of the lecture content. Lecture audio is transcribed and refined into coherent instructional text, which is then used both to generate avatar-delivered lecture videos and to build a temporally aligned textual knowledge base. The lecture content is segmented into timestamped units that serve as the foundation for content-aware retrieval during interactive question answering. Detailed descriptions of this pipeline are provided in Section~\ref{sec:lectureprep}.

\subsection{Stage 2: Interactive Question Answering}

The second stage enables real-time interaction during lecture viewing. When a student pauses the lecture, ALIVE accepts questions via text or voice. The system then performs \emph{content-aware retrieval} by combining semantic relevance with timestamp alignment, ensuring that retrieved lecture segments are both topically related to the question and temporally aligned with the moment of inquiry. These retrieved segments are used to construct a grounded prompt for a locally hosted language model, which generates a concise, contextually consistent explanation. This retrieval and reasoning process is described in detail in Section~\ref{sec:qa}.

\subsection{Stage 3: Avatar-Delivered Response Generation}

The third stage delivers generated explanations in an engaging, instructor-style format. In addition to textual responses, ALIVE can synthesize short talking-head avatar videos that verbally present the explanation. To maintain responsiveness, longer explanations are automatically segmented and played sequentially with progressive preloading. This design reduces perceived latency while preserving visual continuity and a natural instructional flow. The avatar synthesis and playback mechanism is detailed in Section~\ref{sec:avatar}.

\subsection{Frontend--Backend Interaction}

The frontend and backend communicate through lightweight, structured interfaces. The frontend is responsible for detecting user interactions, capturing queries, and presenting responses, while the backend executes content-aware retrieval, language model inference, and avatar generation. Temporary resources created during interaction are automatically managed and cleaned to ensure efficient local operation. Together, these components form a cohesive pipeline that allows students to request clarification at any moment and receive immediate, content-aware explanations without disrupting the learning flow.

\begin{figure*}[h]
\centering
\begin{tikzpicture}[
    node distance=10mm and 12mm,
    box/.style={
        rectangle, rounded corners, draw, thick,
        align=center, minimum width=27mm, minimum height=10mm,
        fill=green!10
    },
    arrow/.style={->, thick}
]

% ---------- Row 1 ----------
\node[box] (video) {Recorded\\Lecture Video};
\node[box, right=18mm of video] (asr) {ASR\\(OpenAI Whisper)};
\node[box, right=18mm of asr] (llm) {Transcript\\Refinement};

% ---------- Row 2 ----------
\node[box, below=14mm of video] (srt) {SRT\\Parsing};
\node[box, right=18mm of srt] (embed) {Sentence\\Embeddings};
\node[box, right=18mm of embed] (faiss) {FAISS\\Index};

% ---------- Synthesia output ----------
\node[box, below=14mm of llm] (synth) {Avatar Lecture\\(Synthesia)};

% ---------- Arrows ----------
\draw[arrow] (video) -- (asr);
\draw[arrow] (asr) -- (llm);
\draw[arrow] (llm) -- (synth);

\draw[arrow] (video) -- (srt);
\draw[arrow] (srt) -- (embed);
\draw[arrow] (embed) -- (faiss);

\end{tikzpicture}

% \caption{Lecture preparation pipeline used by ALIVE. 
% Recorded lectures are transcribed, refined, rendered as avatar-delivered videos, and converted into timestamp-aligned semantic embeddings for retrieval.}
\caption{Offline lecture preparation pipeline, showing how ASR transcription, transcript refinement, and time-aligned segmentation construct a retrieval-ready lecture index for content-aware interaction.}

\label{fig:prep}
\end{figure*}
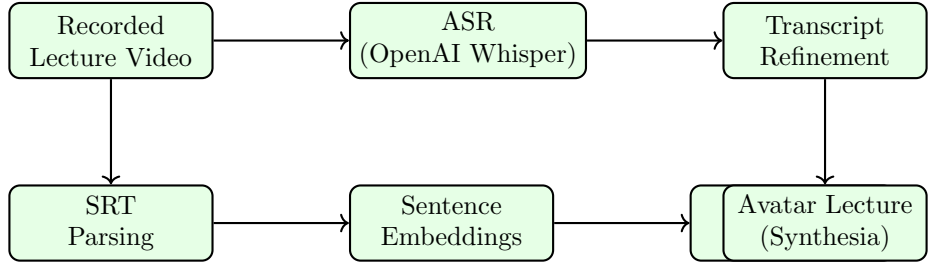

\section{Avatar-Delivered Lecture Preparation}
\label{sec:lectureprep}

The lecture preparation pipeline converts an existing course into an avatar-delivered lecture series and constructs the retrieval-ready knowledge base used for interactive question answering in ALIVE. This stage is executed offline and only once per course, ensuring that all subsequent interactions during lecture playback can be performed efficiently and reliably. As illustrated in Figure~\ref{fig:prep}, the pipeline consists of four main steps: audio transcription, transcript refinement, avatar-delivered lecture synthesis, and retrieval index construction.

\subsection{Audio Extraction and Automatic Speech Recognition}

Each recorded lecture is first converted into an audio track using a standard media processing tool. The audio is then transcribed locally using OpenAI Whisper, an automatic speech recognition (ASR) model capable of producing high-quality transcripts with fine-grained timestamp information. These timestamps associate short spans of spoken content with precise positions in the lecture timeline and later serve as anchors for ALIVE’s content-aware, timestamp-aligned retrieval mechanism.

Performing ASR entirely on local hardware ensures that no lecture audio leaves the institution’s system, supporting privacy preservation and enabling reproducible processing of large lecture archives.

\subsection{Transcript Refinement Using Language Models}

Raw ASR transcripts may contain disfluencies, incomplete sentences, or minor recognition errors. To obtain a coherent and pedagogically clear instructional script, each transcript is refined using a large language model. This refinement step improves grammatical structure and readability while preserving technical terminology and the original instructional intent.

The refined transcript serves two purposes. First, it provides the narration used for avatar-delivered lecture synthesis. Second, it forms a clean and consistent textual source for constructing the retrieval database used during interactive question answering.

\subsection{Avatar-Delivered Lecture Synthesis}

The refined transcript and the original slide deck are rendered into an avatar-delivered lecture using a commercial avatar generation platform, Synthesia \cite{synthesia2024}. In our implementation, a digital avatar modeled after Professor Ge Wang is created and integrated with the refined lecture script and educational slide material. The avatar narrates the refined transcript while the slides advance in synchrony, producing a consistent and polished instructional video.

This process also generates synchronized subtitle (SRT) files that preserve the refined text together with accurate timing information. These subtitle files form the primary textual representation of the lecture content and are subsequently used to construct the retrieval database that supports real-time, content-aware interaction during lecture playback (Section~\ref{sec:qa}).

\subsection{Parsing and Merging Lecture Segments}

The generated SRT files typically contain many short subtitle entries, each spanning only a few seconds. To create a retrieval-friendly structure, ALIVE merges consecutive subtitle entries into longer lecture segments while preserving their temporal alignment. In our implementation, segments are limited to approximately 20 seconds to balance semantic coherence and retrieval granularity.

For each merged lecture segment, the system records its start and end timestamps, the combined transcript text, and a unique segment identifier. These segments form the fundamental units used for semantic and temporal retrieval throughout the system.

\subsection{Embedding Generation and Index Construction}

To support efficient semantic search, each lecture segment is embedded using a lightweight sentence-level transformer model. ALIVE employs an embedding model from the SentenceTransformers family to generate normalized vector representations, which are stored in a FAISS index configured for inner-product similarity search.

For a corpus of $N$ lecture segments with embedding dimension $d$, the index stores an $N \times d$ matrix, enabling fast retrieval even for long lecture series. During interaction, a student’s query is embedded into the same vector space, allowing the system to retrieve semantically relevant segments while maintaining alignment with the lecture timeline through stored timestamps.

\subsection{Resulting Retrieval Store}

The output of the lecture preparation pipeline is a retrieval-augmented generation (RAG) store consisting of: (1) a FAISS index containing segment embeddings, (2) a structured file with timestamped transcript text for each lecture segment, and (3) metadata describing the embedding model and segmentation configuration. This store is loaded by the backend at runtime and serves as the foundation for ALIVE’s content-aware retrieval during interactive question answering.

\subsection{Accessibility and Global Reach}

Beyond enabling interactive retrieval, the lecture preparation pipeline also supports accessibility and global dissemination. Because avatar-delivered synthesis decouples instructional delivery from live recording, the same refined transcript can be re-rendered in multiple languages, facilitating reuse across diverse learner populations. The avatar-delivered lecture videos generated through this pipeline are publicly available via the MILEAGE YouTube channel\footnote{\url{https://www.youtube.com/@MILEAGE-2145}}, providing open access to a complete medical imaging course.

This asynchronous delivery model allows learners across different time zones and educational backgrounds to engage with the material at their own pace. As a result, the lecture preparation pipeline not only enables content-aware interaction but also supports scalable, multilingual, and globally accessible education.

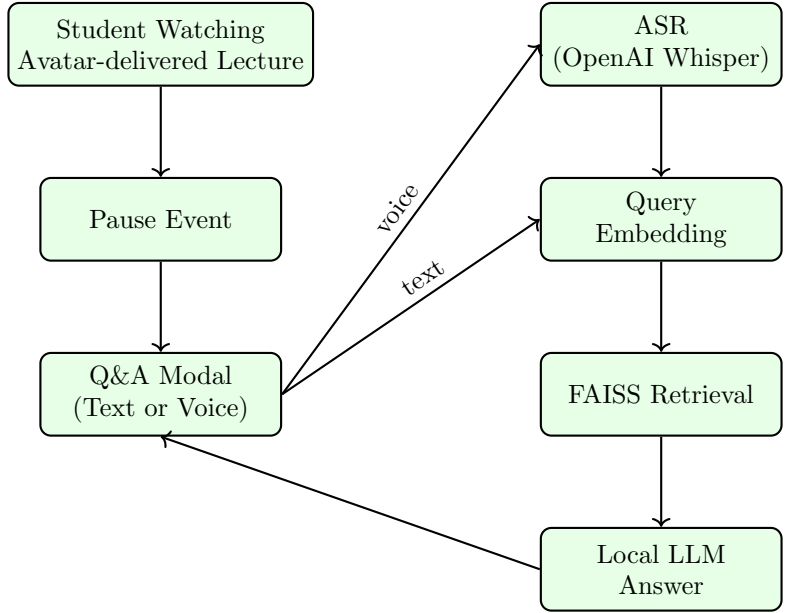
\begin{figure*}[t]
\centering
\begin{tikzpicture}[
    node distance=12mm,
    box/.style={
        rectangle, rounded corners, draw, thick, align=center,
        minimum width=32mm, minimum height=11mm
    },
    arrow/.style={->, thick}
]

% ------------- Left Column (Frontend) -------------
\node[box, fill=green!10] (player) {Student Watching\\Avatar-delivered Lecture};
\node[box, fill=green!10, below=of player] (pause) {Pause Event};
\node[box, fill=green!10, below=of pause] (modal) {Q\&A Modal\\(Text or Voice)};

% ------------- Right Column (Backend) - aligned with player -------------
\node[box, fill=green!10, right=30mm of player] (asr) {ASR\\(OpenAI Whisper)};
\node[box, fill=green!10, below=of asr] (embed) {Query\\Embedding};
\node[box, fill=green!10, below=of embed] (rag) {FAISS Retrieval};
\node[box, fill=green!10, below=of rag] (llm) {Local LLM\\Answer};

% ------------- Arrows -------------
\draw[arrow] (player) -- (pause);
\draw[arrow] (pause) -- (modal);

\draw[arrow] (modal.east) -- (asr.west) node[midway, above, sloped] {voice};
\draw[arrow] (modal.east) -- (embed.west) node[midway, above, sloped] {\qquad text};

\draw[arrow] (asr) -- (embed);
\draw[arrow] (embed) -- (rag);
\draw[arrow] (rag) -- (llm);

\draw[arrow] (llm.west) -- (modal.south);

\end{tikzpicture}

% \caption{Interactive pause-to-question flow in ALIVE. 
% Pausing the lecture triggers the multimodal Q\&A interface, which forwards text or voice queries 
% to the backend. The backend performs timestamp-aware retrieval and generates a grounded response 
% using a local LLM.}
\caption{Pause-triggered, content-aware question answering workflow, demonstrating how semantic similarity and timestamp alignment ground student queries in the relevant lecture context.}

\label{fig:qa_flow}
\end{figure*}

\section{Interactive Question Answering}
\label{sec:qa}

The interactive question answering stage enables students to request clarification at any point during lecture playback and receive explanations grounded in the surrounding instructional content. Unlike conventional chat-based learning tools, this stage is explicitly content-aware: questions are interpreted in the context of the paused lecture segment and answered using content retrieved from the same lecture. This functionality operates entirely at runtime and builds directly on the retrieval store constructed during lecture preparation. This has been shown in Figure \ref{fig:qa_flow}.

\subsection{Pause-Triggered Multimodal Query Interface}

During lecture playback, the frontend continuously monitors user interaction with the video player. When a pause event is detected, the system automatically presents a question-answering modal. It allows students to initiate interaction precisely at the moment confusion arises, without navigating away from the lecture or breaking learning continuity.

The modal supports two input modalities. Students may type a question directly into a chat-style text field or submit a spoken query using the browser’s microphone. The interface remains active until a query is submitted or playback resumes, ensuring minimal disruption to the viewing experience.

\subsection{Voice Query Processing}

For voice-based interaction, the frontend records audio locally using standard web audio APIs. Upon completion of recording, the audio is transmitted to the backend, converted to a suitable waveform format, and transcribed using OpenAI Whisper. If no speech is detected, the system prompts the student to retry; otherwise, the resulting transcription is treated identically to a text-based query and forwarded to the retrieval module.

\subsection{Content-Aware, Timestamp Aligned Retrieval}

Once the question text is obtained—either through direct input or speech transcription—the backend retrieves lecture segments that are both semantically relevant and temporally aligned with the paused lecture position. Let $q$ denote the student’s question and $t$ the timestamp at which the lecture was paused.

Retrieval proceeds in the following steps:
\begin{enumerate}
    \item The question $q$ is embedded using the same sentence-level encoder employed during index construction.
    \item FAISS performs an inner-product similarity search to retrieve the top-$K$ candidate lecture segments.
    \item Each retrieved segment $i$ is associated with a midpoint timestamp $(s_i + e_i)/2$, where $s_i$ and $e_i$ denote the segment’s start and end times.
    \item A temporal adjustment is applied to bias retrieval toward segments near the paused moment:
    \[
        \tilde{d}_i = d_i - \lambda \frac{\left| (s_i + e_i)/2 - t \right|}{60},
    \]
    where $d_i$ is the original semantic similarity score and $\lambda$ controls the strength of temporal sensitivity.
    \item The top-$k$ segments ranked by $\tilde{d}_i$ are selected as contextual evidence.
\end{enumerate}

This content-aware retrieval strategy encourages the system to prioritize instructional material that the student has just encountered, reducing the likelihood of retrieving conceptually distant or temporally irrelevant content.

\subsection{Prompt Construction and Answer Generation}

The retrieved lecture segments are combined with the student’s question to form a structured prompt, which is passed to a locally hosted Llama~3.1~8B language model. The prompt constrains the model to generate explanations grounded in the retrieved lecture context, encouraging consistency with the instructor’s material and terminology.

The resulting answer is returned to the frontend and displayed in the question-answering modal. Students may optionally request the explanation to be delivered as an avatar-delivered response, as described in Section~\ref{sec:avatar}.

\subsection{Interaction Flow and Session Management}

The frontend presents both the student’s question and the generated response in a chat-style interface. All processing—including speech recognition, retrieval, and language model inference—occurs locally, and no interaction data is stored beyond the active session. When lecture playback resumes, the modal automatically closes and the video continues from the paused position.

Through the integration of pause-triggered interaction, multimodal input, content-aware retrieval, and local language model reasoning, ALIVE provides immediate explanations while preserving the natural flow of lecture viewing.

\begin{figure*}[t]
\centering
\begin{tikzpicture}[
    node distance=15mm and 25mm,
    box/.style={
        rectangle, rounded corners, draw, thick, align=center,
        minimum width=30mm, minimum height=12mm,
        fill=green!10
    },
    graybox/.style={
        rectangle, rounded corners, draw, thick, align=center,
        minimum width=30mm, minimum height=12mm,
        fill=green!10
    },
    arrow/.style={->, thick, >=stealth}
]

% ---------------- Row 1: First 3 nodes in sequence ----------------
\node[box] (txt) {LLM\\Text Generation};
\node[box, right=of txt] (audio) {Audio Processing\\TTS + SadTalker };
\node[box, right=of audio] (coeff) {Audio→Coeff\\Extraction};

% ---------------- Row 2: Next 3 nodes in sequence (right-to-left) ----------------
\node[box, below=of coeff] (render) {Neural\\Rendering};
\node[box, left=of render] (segment) {MP4\\Segments};
\node[graybox, left=of segment] (frontend) {Frontend\\Playback};

% ---------------- Arrows: Row 1 left-to-right ----------------
\draw[arrow] (txt) -- (audio);
\draw[arrow] (audio) -- (coeff);

% ---------------- Arrow: Connect row 1 to row 2 (3rd to 4th node) ----------------
\draw[arrow] (coeff) -- (render);

% ---------------- Arrows: Row 2 right-to-left ----------------
\draw[arrow] (render) -- (segment);
\draw[arrow] (segment) -- (frontend);

\end{tikzpicture}

% \caption{Avatar-delivered response generation pipeline in ALIVE.
% The LLM-generated text is converted to speech and preprocessed for SadTalker,
% converted to coefficient representations, rendered into MP4 segments,
% and played sequentially in the browser for smooth real-time delivery.}

\caption{Segmented avatar-delivered response generation pipeline, highlighting how text-to-speech, neural talking-head synthesis, and progressive preloading reduce perceived latency during explanation delivery.}

\label{fig:avatar_pipeline}
\end{figure*}
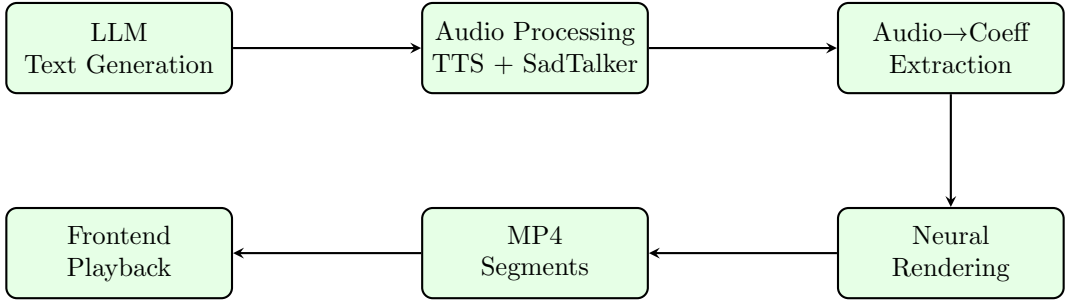

\section{Avatar-Delivered  Response Generation}
\label{sec:avatar}

In addition to textual answers, ALIVE supports an optional avatar-delivered response mode in which explanations are presented verbally through an instructor-style talking-head avatar. This capability is designed to enhance engagement and continuity with the lecture presentation while preserving responsiveness and local execution. Avatar generation is triggered only upon user request and operates as a complementary output modality rather than a mandatory component of question answering. A high-level overview of the avatar-delivered response pipeline is shown in Figure~\ref{fig:avatar_pipeline}.

\subsection{Offline Text-to-Speech Synthesis}

Once the language model generates a textual explanation, the backend converts the text into speech using a fully offline text-to-speech (TTS) engine. The synthesized speech is saved as a temporary WAV file using a unique, timestamp-based identifier to prevent naming conflicts. This audio representation serves as the sole driving signal for subsequent avatar animation and avoids reliance on external TTS services.

\subsection{Talking-Head Animation via SadTalker}

The synthesized speech is passed to a SadTalker-based talking-head animation module, which generates a video of the instructor avatar synchronized to the audio. The animation pipeline consists of three main stages:
\begin{itemize}
    \item \textbf{Audio-driven motion estimation:} Speech features are mapped to facial motion coefficients that drive lip synchronization and local facial dynamics.
    \item \textbf{Reference-guided motion control:} A short reference video provides natural head motion and eye-blink patterns, improving perceptual realism.
    \item \textbf{Neural rendering:} A full-frame talking-head video is synthesized, producing an avatar clip temporally aligned with the spoken explanation.
\end{itemize}

Each avatar request generates a dedicated output directory containing the rendered MP4 file. When necessary, the backend re-encodes the video into a browser-compatible format before returning a playback URL to the frontend.

\subsection{Segmented Synthesis for Long Explanations}

Generating a single avatar clip for long explanations can introduce noticeable startup delays. To reduce perceived latency, ALIVE automatically segments longer answers into shorter units, typically grouped by sentence boundaries, and synthesizes each segment independently.

This segmented synthesis strategy provides two practical benefits:
\begin{itemize}
    \item \textbf{Rapid initial playback:} The first segment becomes available quickly, allowing avatar delivery to begin shortly after answer generation.
    \item \textbf{Background generation:} Remaining segments are synthesized asynchronously while earlier segments are playing.
\end{itemize}

Segmentation decisions are made on the frontend, enabling flexible scheduling without increasing backend complexity.

\subsection{Progressive Preloading and Sequential Playback}

Before avatar playback begins, the frontend preloads the initial video segments to avoid visible stalls. During playback, segment $i$ is displayed while segment $i+2$ is generated or fetched in the background. Playback proceeds sequentially as follows:
\begin{enumerate}
    \item A temporary video element replaces the static avatar portrait.
    \item The animated segment fades in and plays to completion.
    \item The system restores the static portrait and advances to the next segment.
\end{enumerate}

Users may pause or resume avatar playback at any time. All playback orchestration occurs locally within the browser and does not interfere with the main lecture video.

\subsection{Resource Management and Cleanup}

To maintain efficient local execution, all temporary resources generated during avatar synthesis are removed after playback completes. The frontend tracks generated file paths and issues a cleanup request to the backend, which deletes:
\begin{itemize}
    \item Temporary audio files produced during TTS synthesis,
    \item SadTalker-generated MP4 clips, and intermediate re-encoded video files.
\end{itemize}

It ensures that the system remains stateless across interactions and prevents unnecessary accumulation of media files.

\subsection{Integrated Avatar Experience}

By combining local TTS, neural talking-head synthesis, segmented generation, progressive preloading, and automatic cleanup, ALIVE delivers explanations with low perceived latency and consistent visual quality. The avatar appears shortly after answer generation, provides a natural spoken explanation aligned with the lecture content, and seamlessly transitions back to the static portrait once playback concludes. This design enables avatar delivery to function as an engaging and responsive extension of the interactive lecture environment rather than a disruptive or delayed add-on.

\section{Frontend System Design}
\label{sec:frontend}

The frontend provides the user-facing interface through which students watch avatar-delivered lectures, submit questions, and optionally receive avatar-delivered explanations. Implemented as a lightweight web application using standard HTML, CSS, and JavaScript, the frontend is responsible for lecture playback, pause detection, multimodal question input, and the orchestration of segmented avatar responses. Its design prioritizes responsiveness, clarity of interaction, and seamless integration with the lecture flow. Representative interface views are shown in the figures that follow.

% ----------------------------------------------------
% Figure 1: Lecture View
% ----------------------------------------------------
\begin{figure*}[t]
    \centering
    \includegraphics[width=0.7\linewidth]{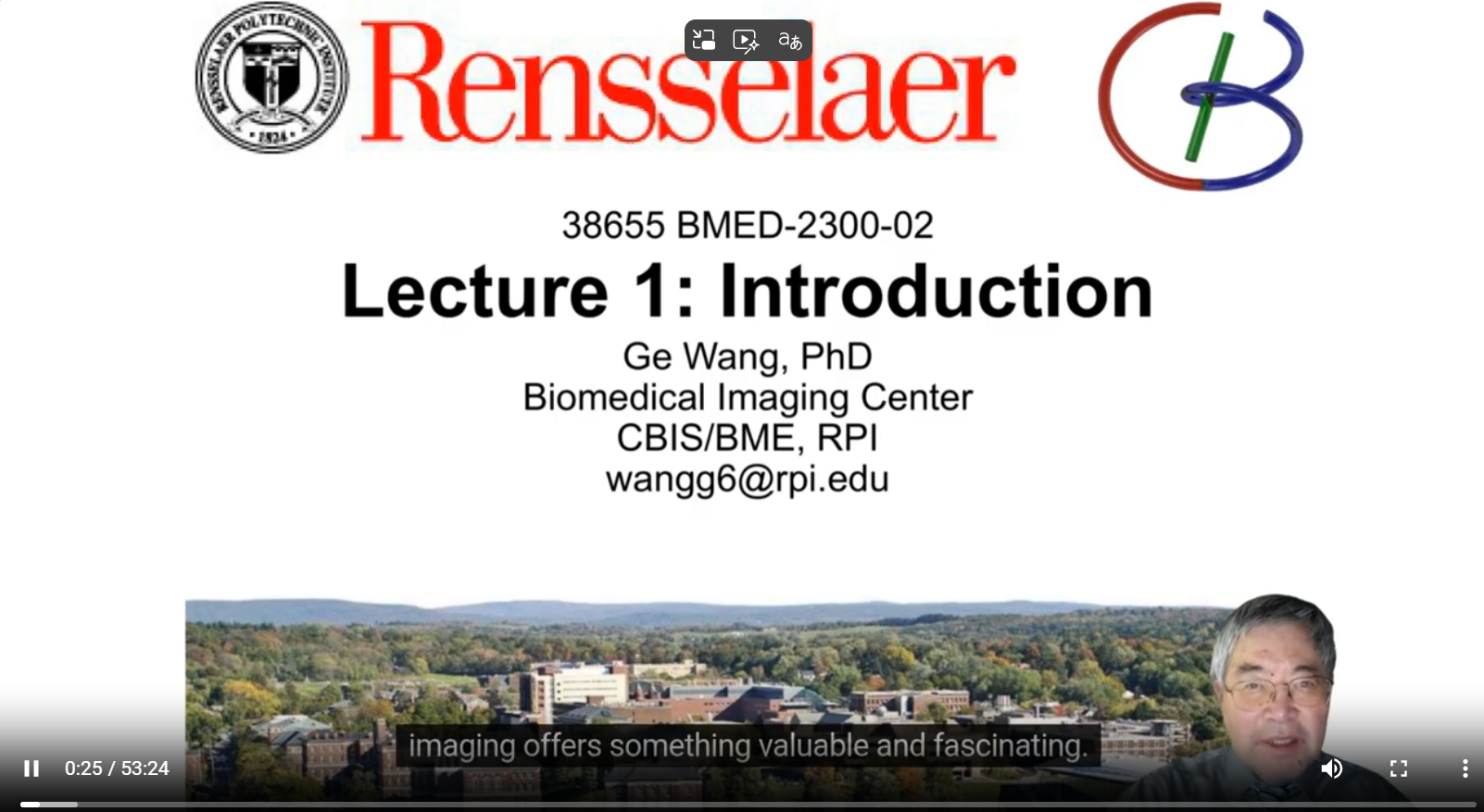}
    % \caption{Avatar-delivered lecture view before any interaction. The lecture plays as a standard video until the student initiates a pause.}
\caption{Baseline avatar-delivered lecture playback interface, providing a consistent instructor presence that anchors subsequent interactive and avatar-delivered explanations.}

    \label{fig:lecture_view}
\end{figure*}

\subsection{Lecture Playback and Interaction}

\begin{figure*}[t]
    \centering
    \includegraphics[width=0.6\linewidth]{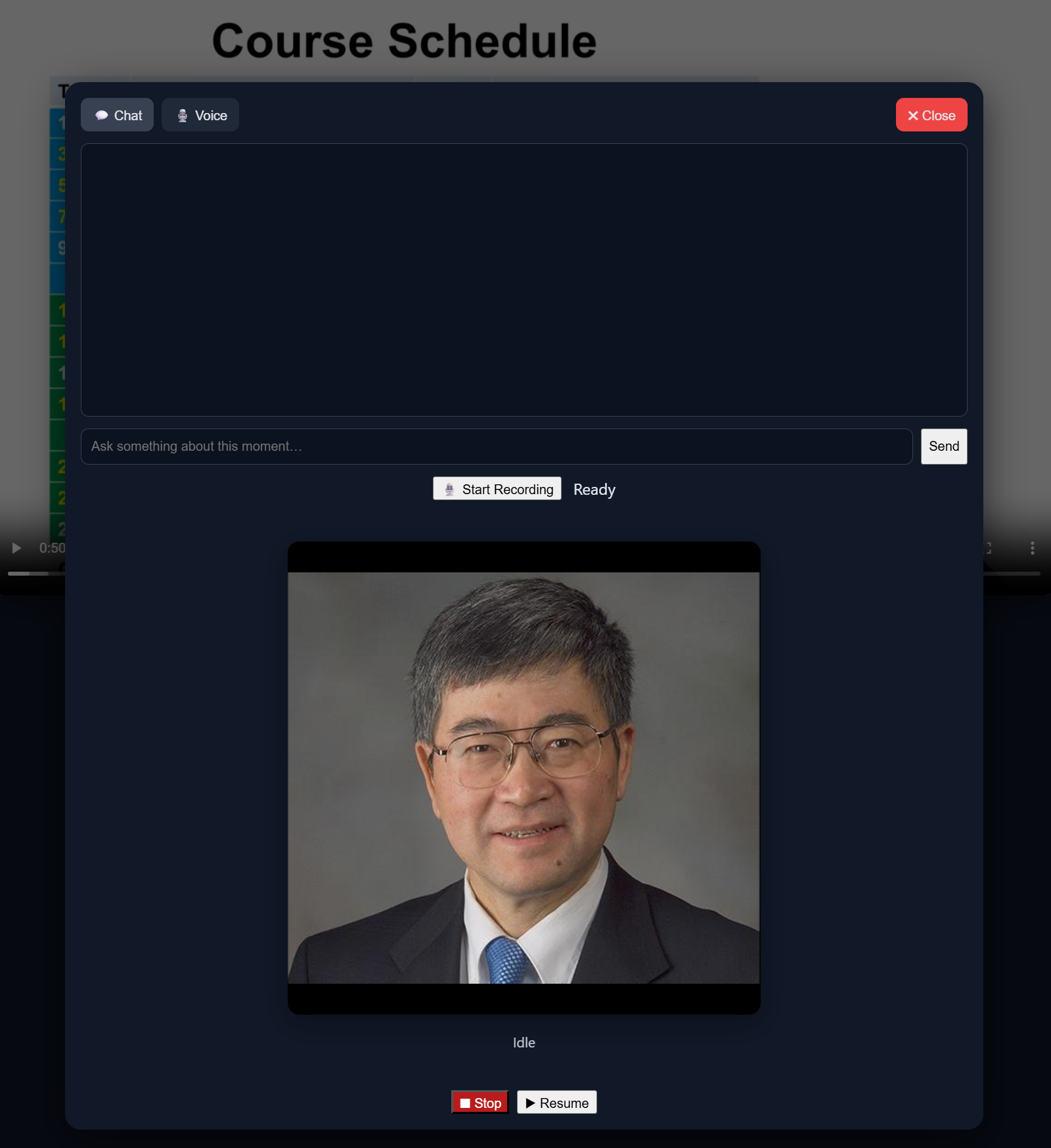}
    % \caption{Pause-triggered Q\&A modal. Students may choose between text-based and voice-based question input.}
\caption{Pause-activated multimodal question interface, enabling students to initiate context-aware interaction precisely at the moment of confusion during lecture viewing.}

    \label{fig:ui_empty}
\end{figure*}

At the core of the frontend is an HTML5 video element that plays the avatar-generated lecture, as shown in Figure~\ref{fig:lecture_view}. The player continuously monitors user interaction events and automatically triggers an interactive question–answering modal when the lecture is paused, illustrated in Figure~\ref{fig:ui_empty}. This design allows students to request clarification precisely at the moment confusion arises, without navigating away from the lecture context.

The modal overlays the lecture video and closes automatically when playback resumes, ensuring a smooth transition between passive viewing and active inquiry.

% ----------------------------------------------------
% Figure 2: Empty Q&A Modal
% ----------------------------------------------------

\subsection{Chat-Based Question Submission}

\begin{figure*}[t]
    \centering
    \includegraphics[width=0.6\linewidth]{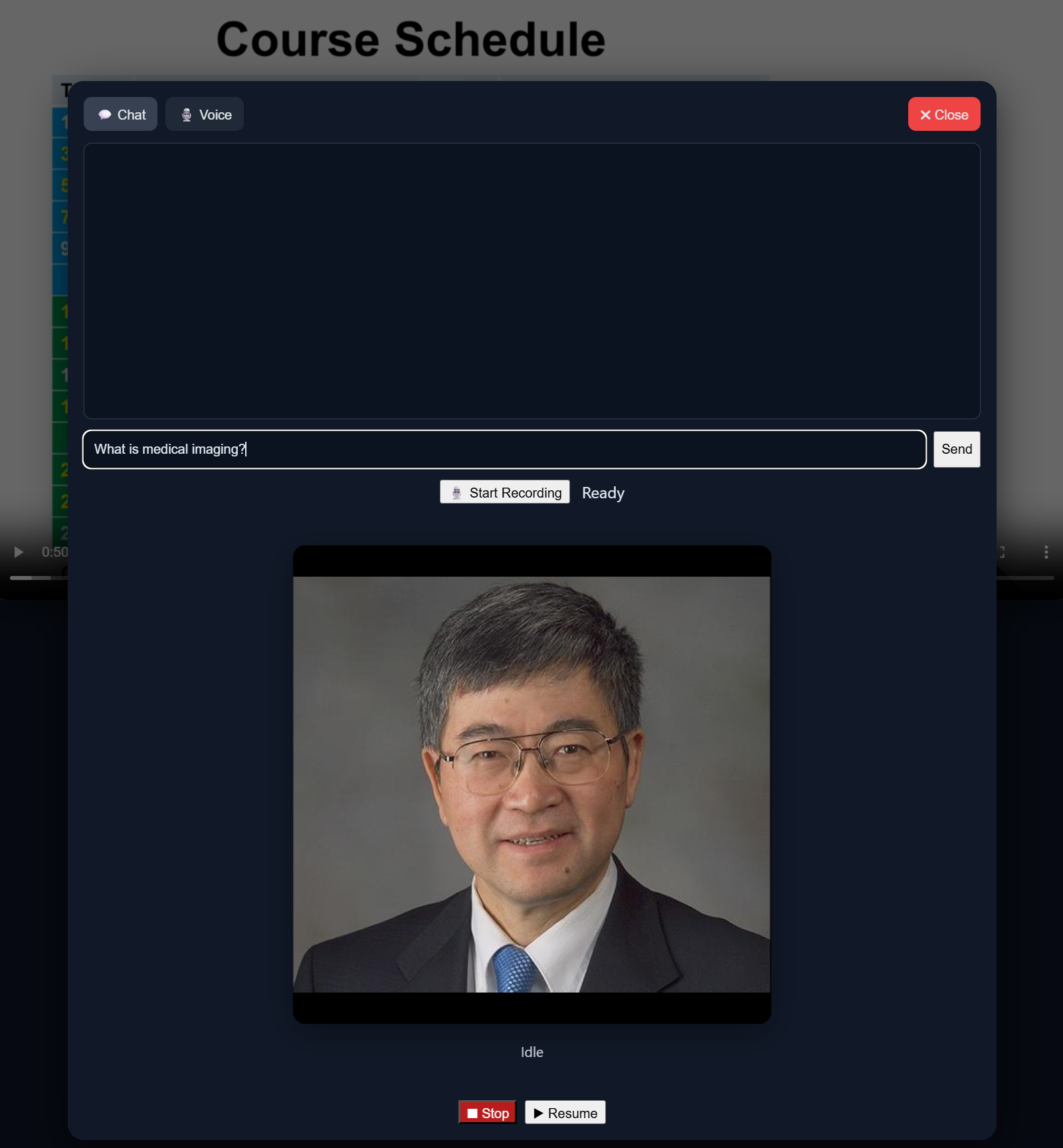}
    % \caption{Text-based question submission within the Q\&A modal (example: ``What is medical imaging?'').}
    \caption{In-context text-based question submission example, illustrating how student queries are embedded within the lecture timeline to support grounded explanation generation (``What is medical imaging?'').}

    \label{fig:ui_question}
\end{figure*}

\begin{figure*}[t]
    \centering
    \includegraphics[width=0.6\linewidth]{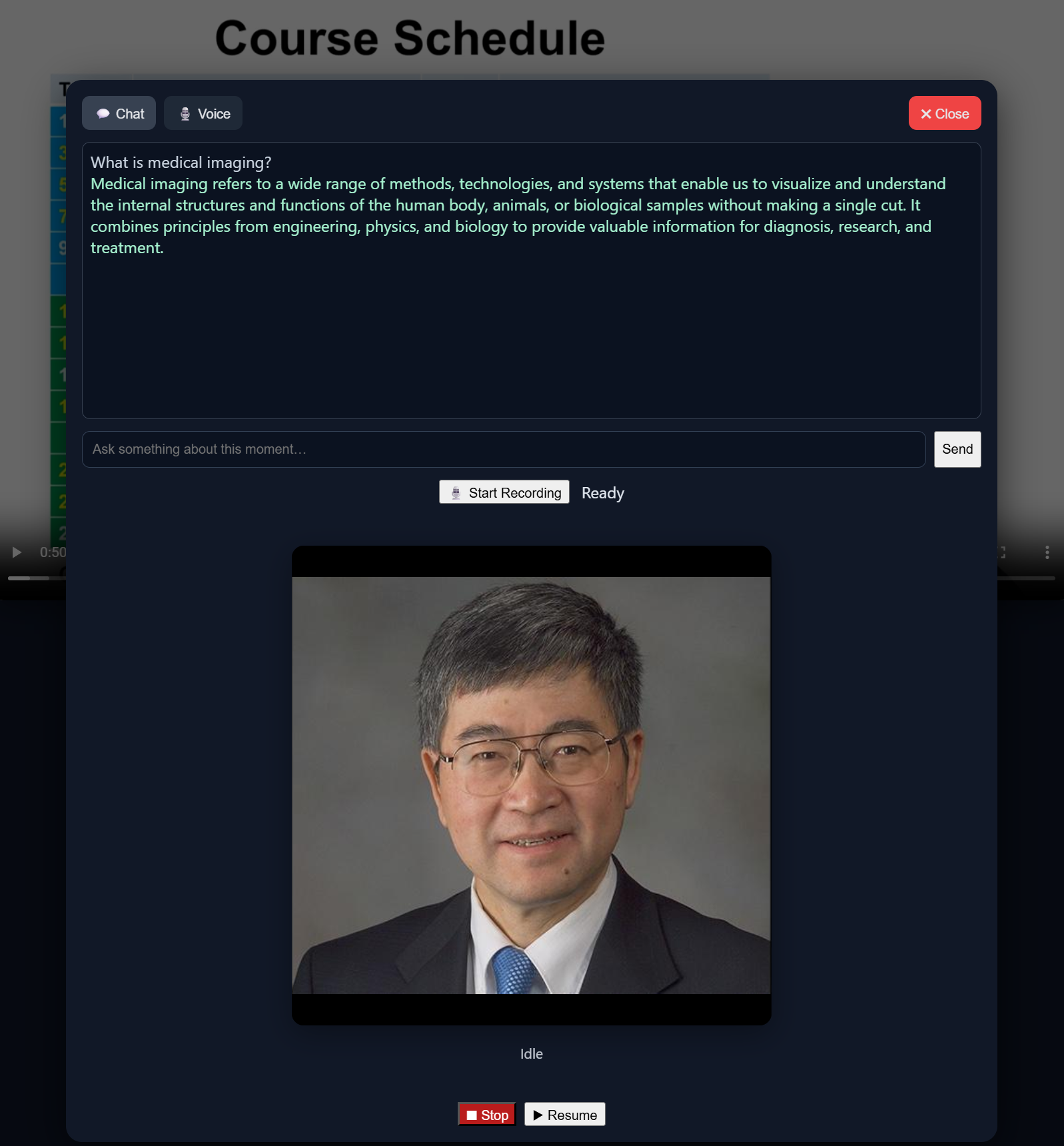}
    % \caption{Grounded textual explanation generated using content-aware retrieval and a local language model.}
\caption{Lecture-grounded textual explanation returned by ALIVE, demonstrating how retrieved lecture segments constrain and guide LLM-generated responses.}

    \label{fig:ui_answer}
\end{figure*}

Within the Q\&A modal, the Chat tab enables students to submit questions through a single-line text input, as illustrated in Figure~\ref{fig:ui_question}. Upon pressing the \emph{Send} button, the query is forwarded to the backend question–answering interface, and both the student’s question and the generated response are appended to a scrollable chat log, providing conversational continuity. Figure~\ref{fig:ui_answer} shows an example of a grounded explanation returned by the system.

% ----------------------------------------------------
% Figure 3: Question UI
% ----------------------------------------------------

% ----------------------------------------------------
% Figure 4: Text Answer UI
% ----------------------------------------------------

\subsection{Voice Input and ASR Feedback}

The Voice tab supports spoken questions using the browser’s built-in audio recording interface. After the student finishes speaking, the recorded audio is transmitted to the backend for transcription using Whisper. Throughout this process, the frontend displays explicit status indicators---such as \emph{Listening}, \emph{Transcribing}, and \emph{Thinking}---to provide transparency about system progress and reduce uncertainty during short processing delays.

If no speech is detected, the interface prompts the student to retry, ensuring robust handling of failed or incomplete recordings.

\subsection{Avatar Container and Interaction States}

A dedicated \texttt{avatar} container manages the visual presentation of instructor-style responses. In its default state, the container displays a static portrait of the instructor. During backend processing, a subtle visual indicator signals that the system is generating a response. When avatar playback is requested, the container transitions to rendering animated talking-head segments generated by the avatar synthesis pipeline.

\begin{figure*}[t]
    \centering
    \includegraphics[width=0.6\linewidth]{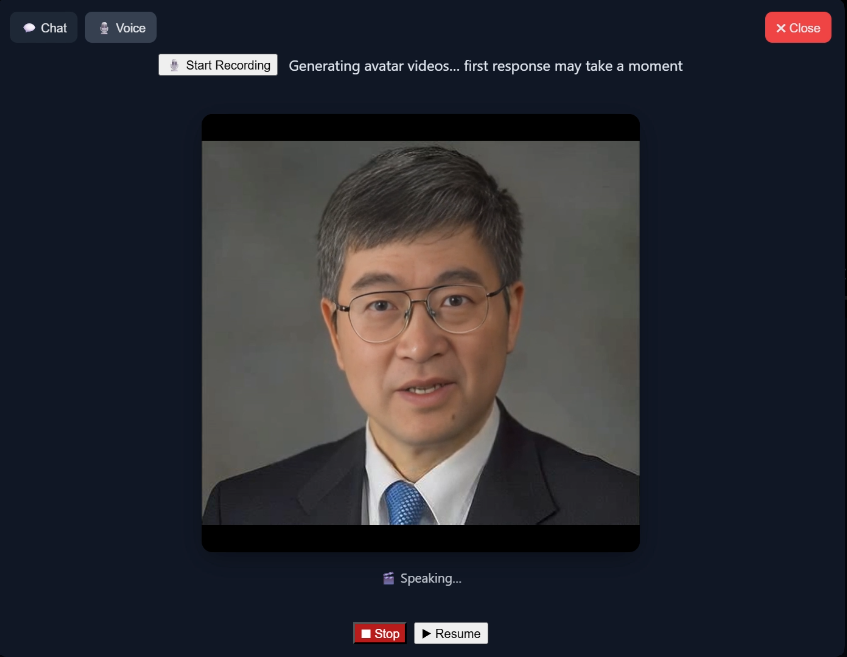}
    % \caption{Instructor-style avatar delivering a spoken explanation through segmented talking-head animation.}
\caption{Avatar-delivered spoken explanation, showing instructor-style verbalization of grounded answers through neural talking-head synthesis.}

    \label{fig:avatar_answer}
\end{figure*}

Figure~\ref{fig:avatar_answer} shows an example of an avatar-delivered explanation rendered within the frontend.

% ----------------------------------------------------
% Figure 5: Avatar Speaking
% ----------------------------------------------------

\subsection{Segmented Avatar Playback and Preloading}

To ensure responsive interaction, the frontend divides longer LLM-generated explanations into sentence-level segments and requests avatar synthesis for each segment individually. Before playback begins, the initial segments are preloaded to minimize startup delay. During playback, segment $i$ is displayed while segment $i+2$ is synthesized or retrieved in the background, enabling smooth and uninterrupted delivery even for multi-paragraph explanations.

This segmentation and preloading strategy significantly reduces perceived latency while preserving visual continuity.

\subsection{Playback Controls and User Autonomy}

The avatar interface provides dedicated \emph{Stop} and \emph{Resume} controls that affect only avatar playback. These controls do not interfere with the lecture video, allowing students to independently manage explanation delivery without disrupting the broader viewing experience.

\subsection{Error Handling and State Reset}

The frontend incorporates safeguards to maintain a stable interaction flow. Empty ASR outputs prompt the student to retry, while failed avatar segments are skipped gracefully to avoid blocking playback. After avatar delivery completes, the frontend initiates cleanup of temporary media files via the backend and automatically resets the avatar container to its idle state, restoring the interface to its default configuration.

\subsection{Lightweight and Local Execution}

The frontend runs entirely within the browser and depends only on standard web technologies. All communication with the backend occurs through JSON-based REST endpoints, and no data is transmitted beyond the local system. This design supports privacy-sensitive deployments and enables reproducible experimentation across different hardware and model configurations.

Overall, the frontend unifies lecture playback, multimodal question submission, and avatar-delivered explanation into a cohesive and intuitive interface that enables real-time, context-aware interaction without disrupting the natural flow of the lecture.

\section{Evaluation}
\label{sec:evaluation}

We evaluate ALIVE along three complementary dimensions: (1) functional correctness, (2) system performance, and (3) user experience. Because ALIVE operates entirely on local hardware and follows a modular design, individual components can be examined in isolation as well as in an end-to-end interactive setting. Rather than benchmarking individual models, our evaluation characterizes the responsiveness, grounding behavior, and interaction behavior of the integrated system during realistic lecture interaction.

\subsection{Functional Evaluation}

\subsubsection{Retrieval Correctness}

To assess content-aware retrieval behavior, we sampled representative student questions aligned with specific lecture moments, such as queries about filtered backprojection during CT reconstruction. For each question, we examined the top-$k$ retrieved lecture segments and evaluated (i) semantic relevance to the question and (ii) temporal proximity to the paused lecture timestamp.

In the majority of cases, retrieved segments were both topically relevant and temporally aligned with the moment of inquiry, indicating that timestamp alignment effectively guided retrieval toward appropriate instructional context. Queries referring to recently presented concepts consistently returned segments from nearby lecture intervals, validating the intended behavior of the content-aware retrieval mechanism.

\subsubsection{LLM Answer Grounding}

We further evaluated whether generated answers remained grounded in the retrieved lecture material. Responses were manually inspected to verify that explanations reflected terminology and concepts present in the retrieved transcript segments and did not introduce unsupported or extraneous information.

Across tested queries, the combination of constrained retrieval and prompt-based grounding resulted in concise answers that closely followed the instructor’s original explanations, demonstrating that ALIVE’s retrieval-augmented pipeline effectively limits hallucination and preserves instructional consistency.

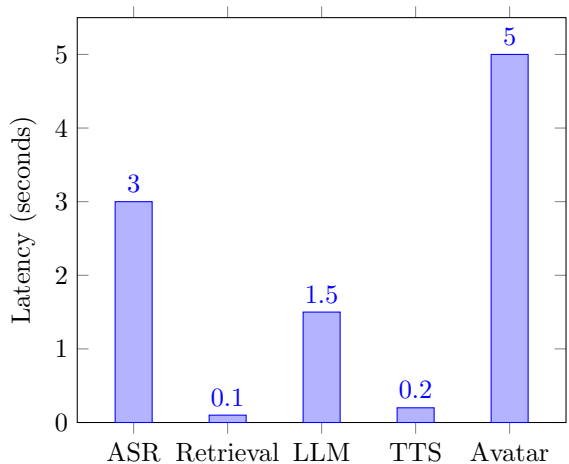
\begin{figure}[t]
\centering
\begin{tikzpicture}
\begin{axis}[
    ybar,
    bar width=14pt,
    width=\linewidth,
    enlarge x limits=0.15,
    ylabel={Latency (seconds)},
    symbolic x coords={ASR,Retrieval,LLM,TTS,Avatar},
    xtick=data,
    ymin=0,
    nodes near coords,
    nodes near coords align={vertical},
]
\addplot coordinates {
    (ASR,3.0)
    (Retrieval,0.1)
    (LLM,1.5)
    (TTS,0.2)
    (Avatar,5.0)
};
\end{axis}
\end{tikzpicture}
% \caption{Representative latency breakdown of major processing stages in ALIVE. Avatar synthesis dominates total response time, while retrieval and text-to-speech remain lightweight. Values are shown as observed ranges during typical interaction.}
\caption{Latency breakdown across ALIVE processing stages, highlighting the relative contributions of ASR, retrieval, LLM inference, and avatar synthesis to end-to-end interaction responsiveness.}

\label{fig:latency}
\end{figure}

\subsection{Performance Evaluation}
\label{sec:performance}

We measured latency across major backend components to assess responsiveness during real-time interaction. All experiments were conducted on the same local workstation used for deployment. Because processing time depends on input length and hardware configuration, we report observed latency ranges during typical usage rather than fixed values.

\paragraph{Automatic Speech Recognition (ASR).}
For spoken queries lasting approximately 5--10 seconds, the CPU-optimized, quantized Whisper model produced transcriptions within roughly 2--4 seconds. As transcription begins only after the user finishes speaking, this delay did not interrupt lecture playback or degrade interaction flow.

\paragraph{Retrieval and Embedding.}
Query embedding and FAISS-based semantic retrieval consistently completed in under 100\,ms for the lecture index size used, contributing negligible latency relative to other components.

\paragraph{LLM Inference.}
The locally hosted Llama-based language model required approximately 1--2 seconds to generate a paragraph-length response. Latency scaled with output length but remained suitable for interactive use during lecture pauses.

\paragraph{Avatar Synthesis.}
Avatar generation dominated overall latency. Text-to-speech synthesis completed in under 0.2 seconds, while SadTalker-based talking-head synthesis required approximately 3--6 seconds per segment depending on hardware availability. To mitigate perceived delay, ALIVE employs segmented avatar synthesis with progressive preloading, allowing the first segment to begin playback while subsequent segments are generated in the background.

Figure~\ref{fig:latency} summarizes the relative latency contributions of major system components.

\subsection{Qualitative Evaluation of Avatar Output}

We qualitatively evaluated the realism and coherence of avatar-delivered responses by examining lip synchronization accuracy, eye-blink behavior, head motion, and visual continuity across segmented playback. Using a short reference video to guide motion synthesis improved perceptual realism and reduced unnatural facial artifacts.

While minor visual discontinuities occasionally appeared at segment boundaries, these artifacts were infrequent and did not noticeably affect comprehension or usability during interaction.

\subsection{Ablation Analysis}

We conducted two lightweight ablations to examine the contribution of key design choices:

\begin{itemize}
    \item \textbf{Without timestamp alignment:} Removing temporal bias during retrieval reduced alignment between questions and lecture context, particularly for short or ambiguous queries.
    \item \textbf{Without segmented avatar synthesis:} Rendering a single long avatar video increased startup latency and negatively affected perceived responsiveness.
\end{itemize}

These observations underscore the importance of content-aware retrieval and segmented avatar synthesis in enabling smooth, real-time interaction during lecture playback.

\section{Discussion}
\label{sec:discussion}

ALIVE demonstrates how recorded lectures can be augmented with real-time, content-aware interaction by tightly integrating content-aware retrieval, locally hosted language models, and neural avatar synthesis within a unified pipeline. Rather than treating question answering, retrieval, and presentation as independent components, ALIVE anchors all interaction around the lecture timeline, enabling explanations that remain aligned with instructional intent and contextual lecture flow. 
This section analyzes the strengths and limitations of ALIVE in comparison with existing systems and discusses the practical implications for interactive learning system design.

\subsection{System Strengths}

A key strength of ALIVE is its ability to deliver \emph{content-aware explanations at the moment of inquiry}. By combining semantic similarity with timestamp alignment, the retrieval mechanism surfaces lecture segments that are not only topically relevant but also temporally consistent with the student’s current viewing position. This design directly addresses a core limitation of existing educational tools by ensuring that generated explanations are grounded in the instructional context in which confusion arises.

Another significant advantage is ALIVE’s fully local execution model. All system components—including automatic speech recognition, retrieval, language model inference, text-to-speech synthesis, and avatar generation—operate on institution-controlled hardware. This design preserves data privacy, avoids reliance on cloud services, and supports reproducible deployment in environments where network access, data governance, or regulatory constraints are critical. The modular architecture further enables individual subsystems to be replaced or upgraded independently without disrupting the overall pipeline.

Finally, avatar-delivered responses provide a consistent instructional presentation modality that complements textual explanations. By maintaining instructor presence during interaction, ALIVE avoids the disjointed experience often associated with external chatbot interfaces. The use of segmented avatar synthesis and progressive preloading allows responses to begin playback promptly, reducing perceived latency while preserving natural delivery.

\subsection{System Limitations}

Despite these strengths, several limitations remain. Avatar synthesis is computationally intensive, and although segmented generation mitigates perceived delay, SadTalker-based rendering still requires several seconds per segment on typical hardware. This constrains responsiveness in resource-limited environments and highlights a tradeoff between visual realism and real-time performance.

Retrieval quality is also sensitive to transcript accuracy and segmentation quality. Errors introduced during automatic speech recognition or transcript refinement may propagate into the retrieval index and affect grounding quality. While timestamp alignment reduces ambiguity for context-specific queries, broad or underspecified questions may still retrieve suboptimal lecture segments.

Additionally, the current retrieval pipeline relies exclusively on subtitle text. Visual instructional elements—such as slides, diagrams, equations, and animations—are not yet incorporated into the retrieval space. Consequently, questions that reference visual material may receive incomplete explanations if relevant information is not explicitly verbalized in the transcript.

Finally, interaction within ALIVE is currently stateless. Each query is processed independently, without dialogue memory or conversational context across turns. While this simplifies system design and reduces complexity, it limits the ability to support extended tutoring interactions or adaptive instructional strategies.

\begin{table*}[t]
\centering
\caption{Comparison of ALIVE with existing systems.}
\label{tab:paper_comparison}
\begin{tabular}{lcccc}
\toprule
\textbf{Method}
& \makecell{\textbf{Lecture}\\\textbf{Aware}}
& \makecell{\textbf{Time-Aware}\\\textbf{Retrieval}}
& \makecell{\textbf{Local}\\\textbf{Pipeline}}
& \makecell{\textbf{Avatar}\\\textbf{Q\&A}} \\
\midrule

OpenAI \cite{openai2023chatgpt}
& $\times$ & $\times$ & $\times$ & $\times$ \\

Johnson \& Lester \cite{johnson2016face}
& $\times$ & $\times$ & $\times$ & \checkmark \\

Zakharov et al. \cite{zakharov2020fast}
& $\times$ & $\times$ & $\times$ & $\times$ \\

Grassal et al. \cite{grassal2022neural}
& $\times$ & $\times$ & $\times$ & $\times$ \\

Su et al. \cite{su2021end}
& $\times$ & $\times$ & $\times$ & $\times$ \\

Xu et al. \cite{xu2023retrieval}
& $\times$ & \checkmark & $\times$ & $\times$ \\

Ren et al. \cite{ren2025videoragretrievalaugmentedgenerationextreme}
& $\times$ & \checkmark & $\times$ & $\times$ \\

Zhao et al. \cite{zhao2025mdkag}
& $\times$ & $\times$ & $\times$ & $\times$ \\

Alawwad et al. \cite{alawwad2025enhancing}
& $\times$ & $\times$ & $\times$ & $\times$ \\

\midrule
\textbf{ALIVE (Ours)}
& \checkmark & \checkmark & \checkmark & \checkmark \\
\bottomrule
\end{tabular}
\end{table*}

\subsection{Comparison with Existing Systems}

Prior work in interactive learning, video question answering, retrieval-augmented language models, and educational avatars addresses important components of the problem, but these capabilities are typically developed in isolation. Text-based large language model assistants and educational retrieval-augmented generation systems provide flexible question answering, yet they are not content-aware and do not align responses with the specific instructional moment at which a learner becomes confused \cite{openai2023chatgpt,alawwad2025enhancing,zhao2025mdkag}. 

Video question answering and retrieval-based video language models extend retrieval to long video content, but they are primarily designed for open-domain video understanding and do not support pause-aligned, timestamp-sensitive interaction during lecture playback \cite{su2021end,xu2023retrieval,ren2025videoragretrievalaugmentedgenerationextreme}. 

Pedagogical avatars and neural talking-head systems improve engagement and visual realism, but they generally rely on scripted content or offline generation and lack real-time, retrieval-grounded question answering tied to lecture context \cite{johnson2016face,zakharov2020fast,grassal2022neural}. In addition, many existing systems depend on cloud-based services, which limits privacy guarantees and reproducibility in educational deployments.

Table~\ref{tab:paper_comparison} summarizes these differences and highlights how ALIVE uniquely integrates content-aware interaction, timestamp-sensitive retrieval, a fully local and privacy-preserving pipeline, and real-time avatar-delivered question answering within a single unified system.

\subsection{Practical and Ethical Considerations}

Local deployment provides strong privacy and reproducibility benefits but introduces practical constraints related to hardware availability. In particular, avatar synthesis performance depends heavily on GPU resources, and CPU-only deployments may experience increased latency for avatar-delivered responses.

Ethical considerations arise from the use of instructor likenesses in talking-head avatars. Although avatars are generated from static images and do not replicate the instructor’s real facial motion, they still convey personal identity and instructional authority. Explicit consent and clear disclosure that explanations are AI-generated are therefore essential. Moreover, the authoritative presentation of avatar-delivered explanations reinforces the importance of strict grounding in retrieved lecture content and cautious handling of uncertainty to avoid over-trust.

\section{Conclusion and Future Work}
\label{sec:conclusion}

This paper presented \textbf{ALIVE}, a fully local, avatar-driven interactive lecture system that augments traditional recorded lectures with content-aware question answering and neural talking-head explanations. By integrating timestamp-aligned retrieval, locally hosted large language models, Whisper-based automatic speech recognition, offline text-to-speech synthesis, and SadTalker-based avatar generation, ALIVE enables students to pause a lecture, pose questions via text or voice, and receive grounded explanations without leaving the instructional context.

System-level evaluation on a complete medical imaging course shows that ALIVE produces lecture-aligned answers with low retrieval latency and acceptable end-to-end response times for interactive use. The use of content-aware retrieval ensures that generated explanations remain consistent with the surrounding lecture context, while segmented avatar synthesis and progressive preloading reduce perceived latency during avatar-delivered responses. These results demonstrate that modern multimodal AI components can be composed into a cohesive, privacy-preserving interactive lecture engine operating entirely on local hardware.

Several directions remain for future work. Incorporating multimodal retrieval over slide images, diagrams, and equations would allow richer grounding for visually oriented questions. Further reductions in avatar synthesis latency—through lightweight or real-time talking-head models and more extensive use of high-performance GPUs—could further enhance system responsiveness. Extending the interaction model to support multi-turn dialogue and contextual follow-up, supported by larger-capacity LLMs, would enable more advanced tutoring-style interactions. Finally, multilingual speech recognition, retrieval, and synthesis could broaden accessibility for global learners.

Overall, ALIVE illustrates how content-aware reasoning, and expressive avatar-delivered presentation can be combined within a fully local deployment model to transform passive lecture videos into responsive, interactive learning environments. This framework provides a foundation for future research on scalable, privacy-conscious, and context-aware educational systems.

\printbibliography

\clearpage

% \appendix
% \section*{Appendix}
% \label{sec:appendix}

% \section{Supplementary Table}
% \label{app:table}

\end{document}